\documentclass{ecai}
\paperid{1174}

\usepackage[pagebackref,breaklinks,colorlinks]{hyperref}
\usepackage{graphicx}
\usepackage{latexsym}
\usepackage{todonotes}
\usepackage{flushend}

\usepackage{amssymb}
\usepackage{subcaption}
\usepackage{booktabs}
\usepackage{algorithm2e}
\usepackage{multirow}

\begin{document}

\begin{frontmatter}

\title{Detecting Out-of-distribution Objects \\ Using Neuron Activation Patterns}

\author[A,B]{\fnms{Bartłomiej}~\snm{Olber}\thanks{Corresponding Author. Email: olberbartlomiej@gmail.com}}
\author[A,B]{\fnms{Krystian}~\snm{Radlak}}
\author[B]{\fnms{Krystian}~\snm{Chachuła}}
\author[A]{\fnms{Jakub}~\snm{Łyskawa}}
\author[A]{\fnms{Piotr}~\snm{Frątczak}}

\address[A]{Warsaw University of Technology, Pl. Politechniki 1, 00-661 Warsaw}
\address[B]{Silesian University of Technology, Akademicka 16, 44-100 Gliwice}



\begin{abstract}
   Object detection is essential to many perception algorithms used in modern robotics applications. Unfortunately, the existing models share a tendency to assign high confidence scores for out-of-distribution (OOD) samples.
Although OOD detection has been extensively studied in recent years by the computer vision (CV) community, most proposed solutions apply only to the image recognition task. Real-world applications such as perception in autonomous vehicles struggle with far more complex challenges than classification. 
In our work, we focus on the prevalent field of object detection, introducing \textbf{N}euron \textbf{A}ctivation \textbf{P}a\textbf{T}te\textbf{R}ns for out-of-distribution samples detection in \textbf{O}bject detectio\textbf{N} (NAPTRON). Performed experiments show that our approach outperforms state-of-the-art methods, without the need to affect in-distribution (ID) performance. By  evaluating the methods in two distinct OOD scenarios and three types of object detectors we have created the largest open-source benchmark for OOD object detection. 
\end{abstract}
\end{frontmatter}

\section{Introduction}
\label{sec:intro}
OOD detection, as described in a thorough study by Yang et al.  \cite{yang2021oodsurvey}, is a problem that arises when machine learning models are applied to data derived from a distribution that is beyond the one they were trained on. 
This results in the model's poor performance since the model had not acquired knowledge that would enable correct predictions. This issue is not limited to classification tasks because object detectors may also encounter OOD samples in two situations: when the inference image contains unknown object classes or when the image scenery is significantly different from the training examples. Practitioners tackle the latter by covering the entire range of scenarios. For example, the most popular autonomous driving datasets include images of environmental, weather, and geographic diversity \cite{nuscenes, bdd_Yu_2020_CVPR, Sun_2020_CVPR}. On the other hand, identifying an unknown object may be possible thanks to algorithms relating to two highly overlapping research fields - open-set (OS) detection \cite{towards_open_set, Dhamija_2020_WACV} and OOD detection \cite{hendrycks17baseline, huang2020survey, vos_du2022towards}. Both OS and OOD methods design a way to quantify the model's uncertainty regarding encountered data so that high uncertainty scores are assigned to unknown objects while low scores are to those well-represented in the training data. 

Contrary to the image classification problem, we observe a severe deficiency of OOD detection methods in object detectors. Detectors process images in a much more complex way than classifiers. Analogously hampered are the evaluation process and posterior analysis of predicted uncertainty scores. This issue became a substantial obstacle for researchers to invent novel OOD algorithms for object detection. Finally, almost every published OS or OOD algorithm assumes Faster R-CNN as the default architecture, making the proposed solution architecture-specific, and unfeasible to apply to any other model. We notice the need for a universal, simple, and yet efficacious OOD framework, which we address in this work.

In this work, we present NAPTRON - i.e., a neuron activation pattern (NAP) OOD method adapted to object detection. It was proved that NAP is a highly efficient technique for OOD detection in image recognition problems \cite{Olber2023}. Our algorithm leverages object detectors' internal feature representation and enables an understanding of training distribution to estimate the uncertainty of predicted bounding boxes.
ReLU-activated layers, being the foundation of most network architectures \cite{He2015DeepRL,redmon2016you,retinanet_Lin_2017_ICCV,carion2020end}, are the natural source of binary patterns, because they set every neuron (or convolution unit) in either positive i.e., on- or zeroed i.e., off-state. The NAPs of ReLU networks display a very convenient property for OOD detection; namely, ReLU-activated networks generate much fewer nonidentical NAPs than they are theoretically capable of generating \cite{HaninRolnick2019,Hartmann2021}. This finding fuels the intuition that if one memorized all known patterns, which are not very numerous, then encountering an unseen pattern during inference would be a reliable indicator of OOD data.

The main contributions of this paper are:
\begin{itemize}
\item We present a theoretically inspired NAPTRON that uses binary NAPs extracted from hidden layers of object detectors to tell ID from OOD predictions. This method is both computationally efficient and practically effective for OOD detection, making it simple to incorporate into existing object detection architectures.
\footnote{\href{https://github.com/safednn-group/naptron}{https://github.com/safednn-group/naptron}}
\item We perform comprehensive experiments involving two datasets and three network architectures, which prove that the proposed method outperforms state-of-the-art OOD detectors.

\item We introduce a novel OOD detection evaluation protocol that analyzes scores of OOD bounding boxes and allows for a more objective comparison of the methods.
    
\end{itemize}

\section{Related Work}\label{related-work}
\subsection{Out-of-Distribution Detection}
Heyndrycks et al. \cite{hendrycks17baseline} proposed a baseline OOD samples detection method. It only involves setting a threshold on the winning class softmax probability (maximum softmax probability, MSP). In object detection, it is already necessary to set a softmax probability threshold for non-maximum-suppression (NMS) to filter redundant background predictions. Therefore, setting another one to separate OOD objects can not be particularly useful in practice. Nevertheless, it still provides a decent baseline.

Initially, Helmholtz \textbf{Energy} \cite{liu2020energy}  had been successfully utilized in OOD detection in image classification. It is a very easy-to-use and universal way to estimate semantic uncertainty. It does not require any detector customization or loss handcrafting. Internally, object detectors perform the classification of proposals resulting in a classification vector for which one can compute the energy score. 

\textbf{Virtual outlier synthesis} (VOS) \cite{vos_du2022towards} is the first work focusing on OOD identification in the object detection task. VOS allows OOD detection by synthesizing virtual outliers in feature (latent) space, thereby regularizing the model’s decision boundary during training. VOS samples the virtual outliers from the low-likelihood region in the feature space and uses them as input for an unknown-aware training objective. The contrastive loss function shapes the uncertainty (Helmholtz energy) space between known data and synthesized outlier data. VOS is not architecture-agnostic because the outlier synthesis process occurs in feature space bound to the fully-connected layers of Faster R-CNN ROI head.  The authors of VOS proposed an evaluation scheme that requires an OOD dataset i.e. a set of images that do not contain ID categories. Any output bounding box generated for these images is considered an "OOD object", while any bounding box generated for images from the ID test dataset is deemed an "ID object". We find this approach is over-simplified because model predictions are not evaluated, so any "OOD object" or "ID object" can actually be a background prediction.

\subsection{Open-Set Detection}
\textbf{Gaussian mixture models} (GMM) \cite{gmm_miller2021uncertainty} approach also introduces a change in the default loss function, focusing on classification loss. The authors add an anchor loss term to facilitate
learning a structured logit space. Next, they fit class-specific GMMs to the logit space with a validation dataset. For any test sample, uncertainty is estimated as a log-likelihood of belonging to any one of the known GMMs.

The authors of \textbf{OpenDet} \cite{han2022opendet} follow the intuition that known-class objects tend to be clustered to form high-density regions in the latent space, while unknown objects are distributed in low-density regions. Consequently, they propose identifying unknown objects by separating high- and low-density regions in the latent space using a contrastive loss. On top of this, they provide another loss function component responsible for learning predictive uncertainty directly as a softmax probability without the logit of ground-truth class. The authors of OpenDet provide the implementation for both Faster R-CNN and RetinaNet. However, the method does not apply to anchor-free architectures because IOU-based sampling of proposals for contrastive learning is not feasible.

The \textbf{OWOD} \cite{owod_joseph2021open} approach was developed before OpenDet. Both ideas share the part of contrastive clustering in latent space. The authors of OWOD also propose an unknown-aware RPN head and the Helmholtz energy-based unknown identification. They model the energy distribution of the known and unknown energy values with a set of shifted Weibull distributions. Similarly to GMM, they fit the distributions using a validation dataset. Due to the RPN requirement, Faster R-CNN is the only compatible architecture.   

\subsection{Other uncertainty estimators in Object Detection}

Previous work by Choi et. al. \cite{Choi_2019_ICCV} introduced \textbf{Gaussian} YOLOv3. The authors redesign YOLO's loss function to model localization uncertainty directly. Each coordinate of a bounding box is represented by a pair of Gaussian parameters (mean and variance) instead of just one value. There is no ground truth for the variance of coordinates, but Gaussian negative-log-likelihood loss conveniently requires ground truth for the mean only, which allows one to learn variance without providing the ground truth. We managed to extend this approach to Faster R-CNN and RetinaNet architectures. In FCOS, however, a "localization centerness" branch is designed to perform a very similar role as Gaussian localization uncertainty; trying to combine them both led to poor results.
    
The popular \textbf{Monte-Carlo Dropout} (MCD) \cite{gal2016dropout} method has already been adapted to the object detection task \cite{Miller2019EvaluatingMS, Miller_2019_CVPR_Workshops}. The most straightforward adaptation technique was proposed in ref. \cite{Miller_2019_CVPR_Workshops}, in which the authors suggested averaging both classification and regression vectors over the number of Monte-Carlo forward passes for each proposal separately. The variance of these vectors for each proposal is the final uncertainty measure.
    
Faster R-CNN-based \textbf{Object Localization Network} (OLN) \cite{kim2021oln} learns to detect objects focusing on localization instead of foreground-background classification. The authors argued that focusing on learning "objectness cues", for the price of obtaining class-agnostic outputs is the best way to achieve cross-dataset generalization. In practice, their method requires replacing the default classification losses in both Faster R-CNN stages with a pair of centerness and IOU losses.

As an alternative baseline focused on objectness for Faster R-CNN, we attempt to establish a threshold for the RPN objectness score of a bounding box proposal. This score is later converted into the final model output. Although the RPN objectness score is not used in the ROI head, we aim to determine its potential usefulness for the OOD detection tasks.

\section{Problem Setup}
Let us denote the set $C$ of ID classes as $K_{ID} = \{1, \:.\;.\;.\;, C\}$ and given object detection dataset $D = \{X,\; \Upsilon\}$ where $X$ and $\Upsilon$ denote the input images and labels. The set of labeled images consists of $N$ samples, $X = \{x_1, \:.\;.\;.\; , x_N\}$ along with the labels associated with the sets of objects included in each of N images $\Upsilon = \{Y_1, \:.\;.\;.\; , Y_N \}$. Next, for each image, we have  $Y_i = \{y_1, \:.\;.\;.\;, y_K \}$ that represents a set of labels for $K$ object instances. Each instance $y = [c,\; l]$ consists of class labels $c \in K_{ID}$ and locations $l = [x,\; y,\; w,\; h]$, where $x, y, w,$ and $h$ denote the object's bounding box center coordinates and size.

An OOD dataset is defined identically, except for the fact that it is semantically 
disparate with the ID dataset, meaning there are no common classes between ID and OOD datasets - $K_{OOD} = \{C + 1, \:.\;.\;.\;\}$. 

Let us also introduce the test dataset $D_{test}$ consisting of ID and OOD objects and an object detector $\theta : x \mapsto Y $ trained on the train split ($D_{train}$) of $D$. 

The OOD detection task is performed for $\theta$-predicted ID and OOD $ \hat{\Upsilon}_{\theta} = \{\hat{\Upsilon}_{ID}, \;\hat{\Upsilon}_{OOD}\}$ samples where ID samples are true positive predictions $\hat{\Upsilon}_{ID} := \{ \hat{y}_{i} \in K_{ID} \times \mathbb{R}^4 : \exists y_t \in D_{test}, \: IOU( \hat{y}_{i}, \, y_t) > \lambda,\: \hat{c} = c_t$ \}. The notation states that true positive predictions are all pairs of class labels and bounding box locations that sufficiently overlap with any ground truth test bounding box of the same class.
Analogously, OOD samples are those predictions, that sufficiently overlap with any ground truth bounding box of unknown class  $\hat{\Upsilon}_{OOD} := \{ \hat{y}_{o} \in K_{ID} \times \mathbb{R}^4 : \exists y_t \in D_{test}, \: IOU( \hat{y}_{o}, \, y_t) > \lambda, \: c_t \in K_{OOD}$ \}. The IOU threshold $\lambda$ is typically set to 0.5.
Now, the goal of an OOD detector is to perform binary classification of $\hat{\Upsilon}_{\theta}$, which relies on the uncertainty estimator $\Phi_{\theta} : \hat{y} \mapsto \phi $ that is intended to assign high scores to OOD samples and low ones to ID samples. Next, the classification is performed by comparing a predefined uncertainty threshold with the uncertainty score associated with each sample.


\section{Our approach}
\subsection{Neuron Activation Patterns}
NAPs have been previously used for uncertainty estimation in image classifiers \cite{cheng2019runtime, Olber2023}. 
Image processing neural networks (NNs) use multiple layers that transform input samples sequentially into the desired form. Nonlinear activation functions present in each hidden layer of a NN significantly contribute to the network's ability to approximate complex, multidimensional relationships. Of all activation functions, ReLU is nowadays the most common choice for CV problems. All ReLU-activated layers produce matrices of either positive or zeroed values, which correspond respectively to active and inactive layers' units. An activation pattern is obtained by assigning a \textit{true} to positive units and \textit{false} to zeros. Thus, for a given network layer, NAP is a binary interpretation of which neurons or convolution units of the layer were activated during the processing of an image.

\subsection{Uncertainty estimation}

Algorithm \ref{alg:nap} describes how NAPTRON estimates the uncertainty of predicted bounding boxes. 
First, for every training image, we perform an object detection inference, simultaneously extracting binary NAPs corresponding to output bounding boxes. The patterns are extracted from a pre-selected layer. Then, for every true positive prediction, we store the extracted pattern in a memory structure. Each object class has a dedicated memory instance. For a given test sample we perform the inference and NAP extraction again. Next, for each inferred bounding box, we find the Hamming distance between its NAP and the nearest pattern out of those stored in the memory structure corresponding to the predicted label. The Hamming distance to the nearest known NAP is the NAPTRON uncertainty estimate. 
 
The authors of \cite{Olber2023} provided an efficient implementation of finding the minimal Hamming distance to known NAPs, which makes real-time uncertainty estimation possible. In this implementation, the Hamming distances between the test NAP and all known NAPs are computed concurrently. The minimal distance is the output of the uncertainty estimation algorithm.

\begin{algorithm}
\caption{NAPTRON uncertainty estimation}\label{alg:nap}
\KwData{$\theta$, $D_{train}$, $x_{test}$, Layer index $l$}
\KwResult{$\phi_{test}$}

\For{$c$ in $K_{ID}$}{
    \tcp{initialize a data structure for each class} 
    $M_{c} \leftarrow \emptyset $  
}

\For{($x$, $y$) in $D_{train}$}{
    $\hat{Y} \leftarrow \theta(x)$ \;
    $NAPS_{x} \leftarrow extractNAPS(\theta, x, l)$\;
    \For{($\hat{y}$, $NAP$) in ($\hat{Y}$, $NAPS_{x}$)}{
        \If{$\hat{y}$ is a true positive prediction}{
            \tcp{Store NAP in the data structure associated with predicted class $\hat{c}$} 
            $M_{\hat{c}} \leftarrow M_{\hat{c}}\, ||\, NAP$ \;
          }
      }
}
\tcp{test phase}
$\hat{Y} \leftarrow \theta(x_{test})$ \;
$NAPS_{test} \leftarrow extractNAPS(\theta, x_{test}, l)$\;
\For{($\hat{y}$, $NAP$) in ($\hat{Y}$, $NAPS_{test}$)}{
    $NAP_{nn} \leftarrow NearestNeighbour(M_{\hat{c}}, NAP)$\;
    \tcp{Uncertainty of $\hat{y}$} 
    $\phi \leftarrow HammingDistance(NAP, NAP_{nn})$\;
}
\end{algorithm}



\subsection{Pattern extraction for bounding boxes} \label{nap-extraction}
In every state-of-the-art object detector, one can distinguish a detection backbone part (e.g. ResNet50) that extracts features of an input image and a detection head that performs classification and regression of proposal bounding boxes (priors) based on the extracted features.
Typically, the classification and regression are computed in two separate subnetworks that work in parallel. The \textit{extractNAPS} function extracts binary NAPs of the $l$-th ReLU-activated layer of the classification branch in the detection head of object detector $\theta$. For Faster R-CNN, the ROI-head consists of fully connected (FC) layers. The operation is straightforward because there is no ambiguity in the link of hidden layers' activations to the final output. Assuming $P$ proposals generated for a single image by the RPN-head, each layer processes $[P, In]$-shaped feature maps into $[P, Out]$-shaped feature maps, where  $In$ is the number of neurons of the previous layer and $Out$ is the number of neurons of the current layer. Thus, for each layer and each proposed bounding box, we can easily distinguish an activation pattern of length $Out$.

Nevertheless, detection heads of single-stage architectures such as RetinaNet and FCOS, use convolutional layers instead of FC layers. Each hidden layer takes as an input $[W, H, C]$-shaped feature maps and processes them without changing their dimensions. The final layer changes the number of channels of the matrix from $C$ to $K*A$, where $W$ and $H$ are the width and height of processed feature maps, respectively, $K$ is the number of classes, and $A$ is the number of priors with centers in each of $W * H$ locations of the feature maps. In our approach, all the $A$ possible bounding boxes predicted in the $(w, h)$ location are associated with a $C$-long activation vector located at $(w, h)$ of the chosen hidden layer's output. In other words, any $(w, h)$-centered output bounding box is attributed to the activation values located in the same coordinates of the hidden feature maps.

\section{Experiments} 

We propose an evaluation protocol to examine the effectiveness of the proposed NAPTRON detector in various aspects. These include the ability to identify OOD objects, recognize known from unknown samples, and gradually acquire knowledge about new categories when labels are available for certain unknown objects.

 \noindent {\bf Datasets.} The experiments were conducted in two domain shift scenarios.  In the first scenario, the detector was trained on the train split of the PASCAL VOC \cite{pascal_voc_Everingham10} dataset consisting of 20 classes. Evaluation protocol for known and unknown objects was performed on validation split of the COCO \cite{coco_Lin2014MicrosoftCC} dataset consisting of 80 classes, including the 20 known ones. The validation split includes 20631 known objects and 15704 unknown objects. In the second scenario, the detector was trained on images from the train split of the BDD100k \cite{bdd_Yu_2020_CVPR} dataset that consists of only 4 known classes (i.e., "pedestrian", "bicycle", "car" and "traffic sign"). Evaluation protocol was performed on the full validation split of the BDD100k dataset consisting of all 10 classes, including the 4 known ones. The validation split includes 152025 known objects and 33920 unknown objects.

 \noindent {\bf Architectures.} For our experiments, we chose three well-established object detection architectures: Faster R-CNN (two-stage, anchor-based) \cite{faster-rcnn_NIPS2015}, RetinaNet (single-stage, anchor-free) \cite{retinanet_Lin_2017_ICCV},  and FCOS (single-stage, anchor-based) \cite{fcos_Tian_2019_ICCV}. The selected architectures represent different properties that enable our findings to be universal across many state-of-the-art object detection architecture types. All models were trained with default configuration parameters provided in the MMDetection \cite{Chen2019MMDetectionOM} framework.

\subsection{Parameter sensitivity analysis}
The NAPTRON algorithm has a couple of parameters that may affect OOD detection performance quality. We examined in detail the impact of those parameters, i.e.:
\begin{itemize}
    \item layer index,
    \item distance reduction,
    \item binarization percentile threshold $p$,
    \item train samples softmax probability threshold $s$,

    \item NMS softmax score threshold.
\end{itemize}
\noindent {\bf Layer index.} Choosing an optimal layer to extract binary NAPs from is very difficult when detecting OOD samples in the image classification task \cite{Olber2023} since modern classifiers consist of a large number of layers. Detection heads of standard architectures - such as  Faster R-CNN, RetinaNet, and FCOS - have only a few layers (i.e., 2-4), so choosing the correct one should be much easier. We also try extracting patterns from the $[W, H, C]$-shaped feature maps originating from the feature extracting backbone to the detection head - denoted Layer 0 in Tables \ref{table:Binarization threshold sensitivity}, \ref{table:IOU threshold sensitivity}, and \ref{table:softmax train threshold sensitivity}. For RetinaNet and FCOS, extracting NAPs from Layer 0 requires identical operations as extracting from other layers, as described in Section \ref{nap-extraction}. However, for Faster R-CNN, we had to flatten the feature maps to match the dimensionality of the FC layers. To do so, we computed the $C$-long vector of the means of every channel and binarized it by zeroing $p$ percent of the lowest values in the vector.    

\noindent {\bf Distance reduction.} The authors of \cite{Olber2023} estimate uncertainty by finding the minimal Hamming distance between the test NAP and all known NAPs of the predicted class; we check whether computing the average distance to the known NAPs yields improved results. 

\noindent {\bf Binarization percentile threshold $p$.} Zeroing a certain percentage of units, which have the lowest magnitude across an activation pattern, was effective in the image classification setup. Our technique of extracting NAPs from detection heads makes the binarization step optional, so the default binarization threshold value equals $0.0$.

\noindent {\bf Training samples IOU threshold.} To construct the database of known patterns, one needs to perform object detection on the training images and choose only those activation patterns that correspond to the true positive predictions. Typically, a prediction is deemed correct if it overlaps any object with IOU greater than 0.5. We want to check if setting a higher IOU threshold (e.g., 0.9), and consequently choosing only the patterns that correspond very accurately to an object might lead to more accurate OOD detection results.

\noindent {\bf Training samples softmax probability threshold $s$.} The reason for examining the effect of this parameter is the same as for the IOU threshold explained above. The default softmax threshold value equals $0.0$, meaning no true positive (TP) sample is discarded. 

\noindent {\bf NMS softmax score threshold.} Every OOD detector is sensitive to this parameter. The impact of the NMS threshold on all OOD methods' performance is described and studied in Section \ref{object-detection-experiments}.

\noindent \textbf{Results.} The experiments were performed on the first evaluation scenario (PASCAL-VOC $\rightarrow$ COCO); AUROC is used as the performance metric. Note that the values of the first columns of each Table \ref{table:Binarization threshold sensitivity}, \ref{table:IOU threshold sensitivity}, and \ref{table:softmax train threshold sensitivity} are identical - all were generated for the default setup of the parameters (IOU $\geq 0.5$, $s \geq 0.0$, $p = 0.0$).

\begin{table}
\scriptsize
\begin{center}
\caption{ Impact of the binarization percentile threshold on AUROC metric using Faster R-CNN and COCO evaluation protocol.}
\label{table:Binarization threshold sensitivity}
\begin{tabular}{llccc} 
\hline\noalign{\smallskip}
Layer index & Distance reduction & p = 0.0 & p = 0.45 & p = 0.9\\
\noalign{\smallskip}
\hline
\multirow{2}{*}{Layer 0} & minimal dist. & 0.7104 & 0.7083 &  0.6946\\
& avg. dist. & 0.7660 & 0.7621 &  0.7426 \\
\midrule
\multirow{2}{*}{Layer 1} & minimal dist. & 0.7302 & 0.7272 &  0.7139\\
& avg. dist. & 0.7328 & 0.7288 &  0.7168 \\
\midrule
\multirow{2}{*}{Layer 2} & minimal dist. & 0.7712 & 0.7671 &  0.7500\\
& avg. dist. & 0.7733 & 0.7694 &  0.7565 \\
\hline
\end{tabular}
\end{center}
\end{table}

\begin{table}
\scriptsize
\begin{center}
\caption{Impact of the NAP train patterns IOU threshold on AUROC metric using Faster R-CNN and COCO evaluation protocol.}
\label{table:IOU threshold sensitivity}
\begin{tabular}{llccc} 
\hline\noalign{\smallskip}
Layer index & Distance reduction & IOU $\geq$ 0.5 & IOU $\geq$ 0.7 & IOU $\geq$ 0.9\\
\noalign{\smallskip}
\hline
\multirow{2}{*}{Layer 0} & minimal dist. & 0.7104 & 0.7106 &  0.7066\\
& avg. dist. & 0.7660 & 0.7658 &  0.7522 \\
\midrule
\multirow{2}{*}{Layer 1} & minimal dist. & 0.7302 & 0.7303 &  0.7371\\
& avg. dist. & 0.7328 & 0.7328 &  0.7475 \\
\midrule
\multirow{2}{*}{Layer 2} & minimal dist. & 0.7712 & 0.7712 &  0.7665\\
& avg. dist. & 0.7733 & 0.7733 &  0.7652 \\
\hline
\end{tabular}
\end{center}
\end{table}

\begin{table}
\scriptsize
\begin{center}
\caption{Impact of the softmax score threshold for training NAPs extraction on AUROC metric using Faster R-CNN and COCO evaluation protocol.}
\label{table:softmax train threshold sensitivity}
\begin{tabular}{llcccc} 
\hline\noalign{\smallskip}
Layer index & Distance reduction & s $\geq$ 0.0 & s $\geq$ 0.3 & s $\geq$ 0.6 & s $\geq$ 0.9\\
\noalign{\smallskip}
\hline
\multirow{2}{*}{Layer 0} & minimal dist. & 0.7104 & 0.7105 &  0.7113  &  0.7158\\
& avg. dist. & 0.7660 & 0.7659 &  0.7651  &  0.7623\\
\midrule
\multirow{2}{*}{Layer 1} & minimal dist. & 0.7302 & 0.7295 &  0.7272  &  0.7242\\
& avg. dist. & 0.7328 & 0.7319 &  0.7309  &  0.7283\\
\midrule
\multirow{2}{*}{Layer 2} & minimal dist. & 0.7712 & 0.7692 &  0.7657  &  0.7576\\
& avg. dist. & 0.7733 & 0.7716 &  0.7702  &  0.7666\\
\hline
\end{tabular}
\end{center}
\end{table}

 Results in Tables \ref{table:Binarization threshold sensitivity}, \ref{table:IOU threshold sensitivity}, and \ref{table:softmax train threshold sensitivity} suggest that tuning training samples' IOU and softmax threshold parameters is not worthwhile. Maintaining the standard 0.5 IOU threshold, the lowest (0.0) softmax threshold - thus letting as many training samples as possible be stored in the known patterns database - allows an obtaining of the optimal results. Moreover, shifts in these parameters do not introduce significant performance variation.
As for the binarization percentile threshold, abstaining from binarization seems to be the best choice. We find that the higher the threshold, the worse the results we obtained. 

On the other hand, introducing the averaging Hamming distance (instead of the previously proposed finding of the distance to the closest binary pattern) brings a slight improvement - especially if Layer 0 is chosen. The choice of the layer is the most important parameter to tune. We recommend extracting patterns from the penultimate layer of the detection head.

\subsection{OOD detection} \label{OOD-detection-experiments}

In the next experiment, we compared the performance of the proposed \hyphenation{NAPTRON} algorithm with the state-of-the-art OOD object detectors:
  \begin{itemize}
      \item confidence score from Faster R-CNN (Standard)
      \item objectness score from Region Proposal Network in Faster R-CNN (RPN)
      \item Energy \cite{liu2020energy}
\item Virtual Outlier Synthesis (VOS) \cite{vos_du2022towards}
\item Gaussian Mixture Models (GMM) \cite{gmm_miller2021uncertainty}
\item OpenDet \cite{han2022opendet}
\item Open World Object Detection  (OWOD) \cite{owod_joseph2021open} 
\item  Gaussian YOLOv3 (Gasussian) \cite{Choi_2019_ICCV}
\item Monte-Carlo Dropout (MCD) \cite{Miller_2019_CVPR_Workshops}
\item Object Localization Network (OLN) \cite{kim2021oln}
  \end{itemize}

  All methods were compared using the officially published code of each algorithm with recommended configuration parameters, except the MCD method, which we implemented ourselves. It is important to notice that some of the methods were only designed for certain DNN architectures (e.g., Faster R-CNN); and we applied the methods only for the dedicated models.

 \noindent {\bf Metrics.}
 We measure OOD detection performance using two metrics: (1) the false positive rate at the true positive rate of 95\% - FPR@95TPR and (2) the area under the receiver operating characteristic curve (AUROC). 
The metrics are computed separately for each class and then averaged.


\noindent {\bf Performance comparison.} Table \ref{table:OOD performance} shows the performance of the proposed algorithm and other methods. As can be observed, NAPTRON achieves the best FPR@95TPR for every evaluation scenario, detector architecture pair considered in this work, and the best AUROC for 3 out of 6 evaluation cases. No other method yields such consistently good results. VOS had been designed specifically for the Faster R-CNN architecture and detects OOD samples roughly just as well as NAPTRON. However, these methods could not be compared using other architectures since there is no straightforward way to apply VOS for anything but Faster R-CNN. 
We suspect that even if it could be applied to the single-stage architectures, it would perform similarly to Energy because both methods rely on the energy score.

Additionally, the results showed that the open-sets methods (OpenDet, GMM, and OWOD) are not the most effective OOD detectors. They fail to outperform the baseline detector (Standard) consistently. GMM manages to do so in 3 or 4 (depending on the metric) out of 6 cases, OpenDet in 1 out of 4, and OWOD in 1 or 2 out of 2.

The rest of the methods perform very poorly in our challenging experimental setup. Gaussian, OLN, RPN, and MCD are unable to beat the baseline detector (Standard).

\begin{table*}[!htb]
\scriptsize
\begin{center}
\caption{Comparison of OOD detectors' performance}
\label{table:OOD performance}
\begin{tabular}{llcc|cc} 
\hline\noalign{\smallskip}
Network & Method & FPR@95TPR $\downarrow$ & AUROC$\uparrow$ & FPR@95TPR$\downarrow$ & AUROC$\uparrow$\\
\noalign{\smallskip}
\hline
\noalign{\smallskip}
& & \multicolumn{2}{c|}{ VOC $\rightarrow$ COCO} & \multicolumn{2}{c}{ BDD100k$\rightarrow$ BDD100k}  \\ \midrule
\multirow{11}{*}{Faster R-CNN} & Energy & 0.6762 & 0.8058 &  0.7987 & 0.7165\\
& Gaussian & 0.8899 & 0.7103 &  0.8458 & 0.6320\\
& GMM & 0.7441 & 0.7881 &  0.8420 & 0.6742\\
& MCD & 0.8267 & 0.6848 &  0.9111 & 0.5778\\
& NAPTRON & \textbf{0.5677} & 0.7786 &  \textbf{0.7603} & 0.6990\\
& OLN & 0.9023 & 0.5994 &  0.9532 & 0.5979\\
& OpenDet & 0.8783 & 0.6965 &  0.9004 & 0.5663 \\
& OWOD & 0.7547 & 0.6858 &  0.8232 & 0.6757\\
& RPN & 0.8784 & 0.6761 &  0.8894 & 0.6220\\
& Standard & 0.8426 & 0.7459 &  0.8358 & 0.6554\\
& VOS & 0.6551 & \textbf{0.8217} &  0.7839 & \textbf{0.7249}\\
\midrule
\multirow{7}{*}{RetinaNet} & Energy & 0.7994 & 0.6865 &  0.9081 & 0.5438\\
& Gaussian & 0.7984 & 0.7456 &  0.8906 & 0.5882\\
& GMM & 0.8044 & 0.7340 &  0.9081 & 0.5925\\
& MCD & 0.8947 & 0.5023 &  0.9482 & 0.5143\\
& NAPTRON & \textbf{0.6063} & \textbf{0.7750} &  \textbf{0.8738} & \textbf{0.6468}\\
& OpenDet & 0.6198 & 0.7717 &  0.9584 & 0.5075\\
& Standard & 0.7917 & 0.7426 &  0.8787 & 0.6041\\
\midrule
\multirow{5}{*}{FCOS} & Energy & 0.8148 & 0.6762 & 0.8778 & 0.5528 \\
& GMM & 0.6985 & 0.7276 &  0.8295 & \textbf{0.7067}\\
& MCD & 0.7721 & 0.5783 &  0.9330 & 0.5095\\
& NAPTRON & \textbf{0.5984} & \textbf{0.7773} &  \textbf{0.7791} & 0.6978\\
& Standard & 0.7721 & 0.7256 &  0.8615 & 0.6334\\
\hline
\end{tabular}
\end{center}
\end{table*}

\begin{figure}[!htb]
\centering
\includegraphics[width=0.9\linewidth]{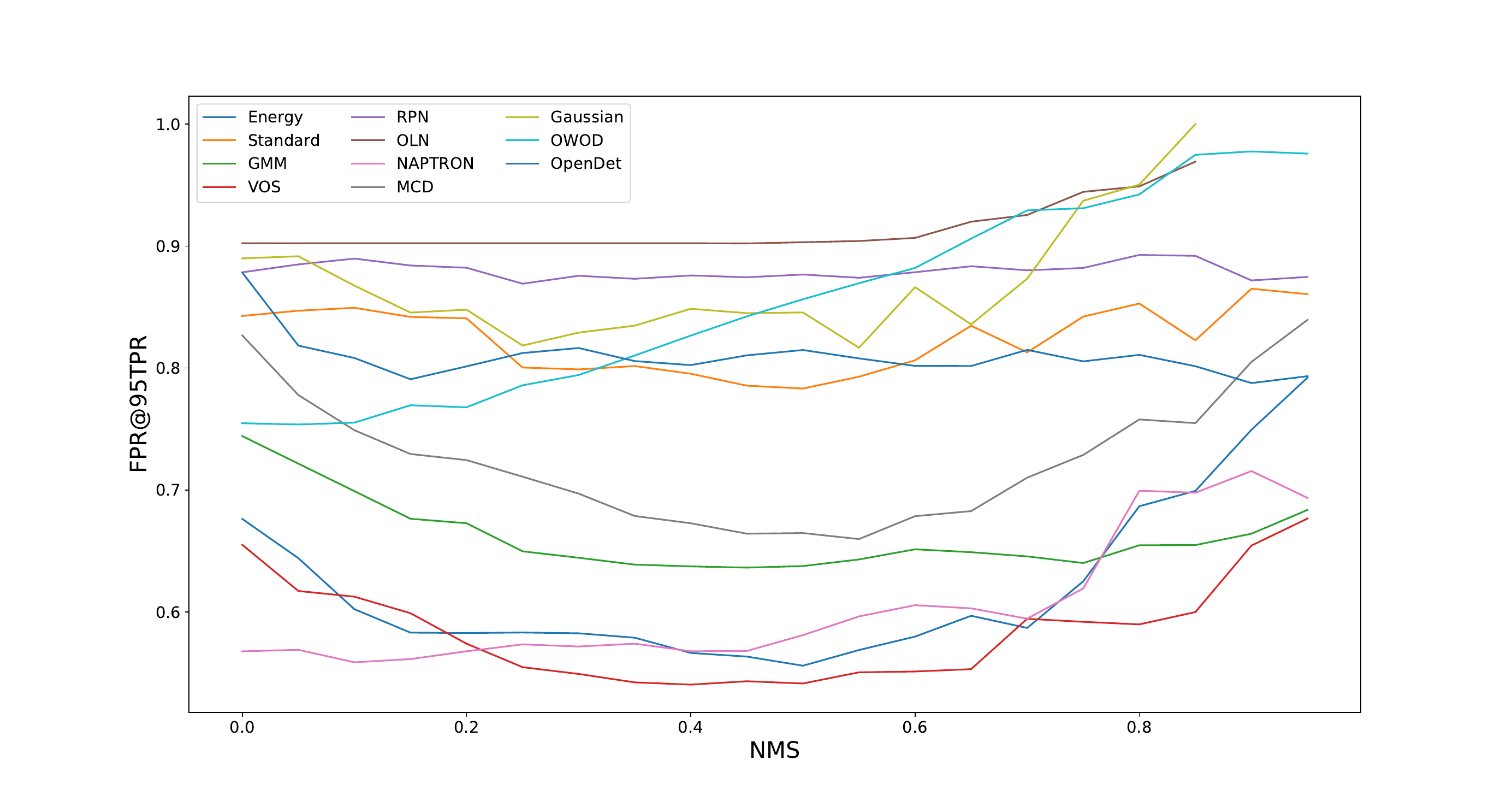}
\caption{Faster R-CNN-based OOD detectors NMS sensitivity evaluated on COCO. Metric - FPR@95TPR.}
\label{fig:frcnn-voc2coco-fpr-sensitivity}
\end{figure}

\begin{figure}[!htb]
\centering
\includegraphics[width=0.9\linewidth]{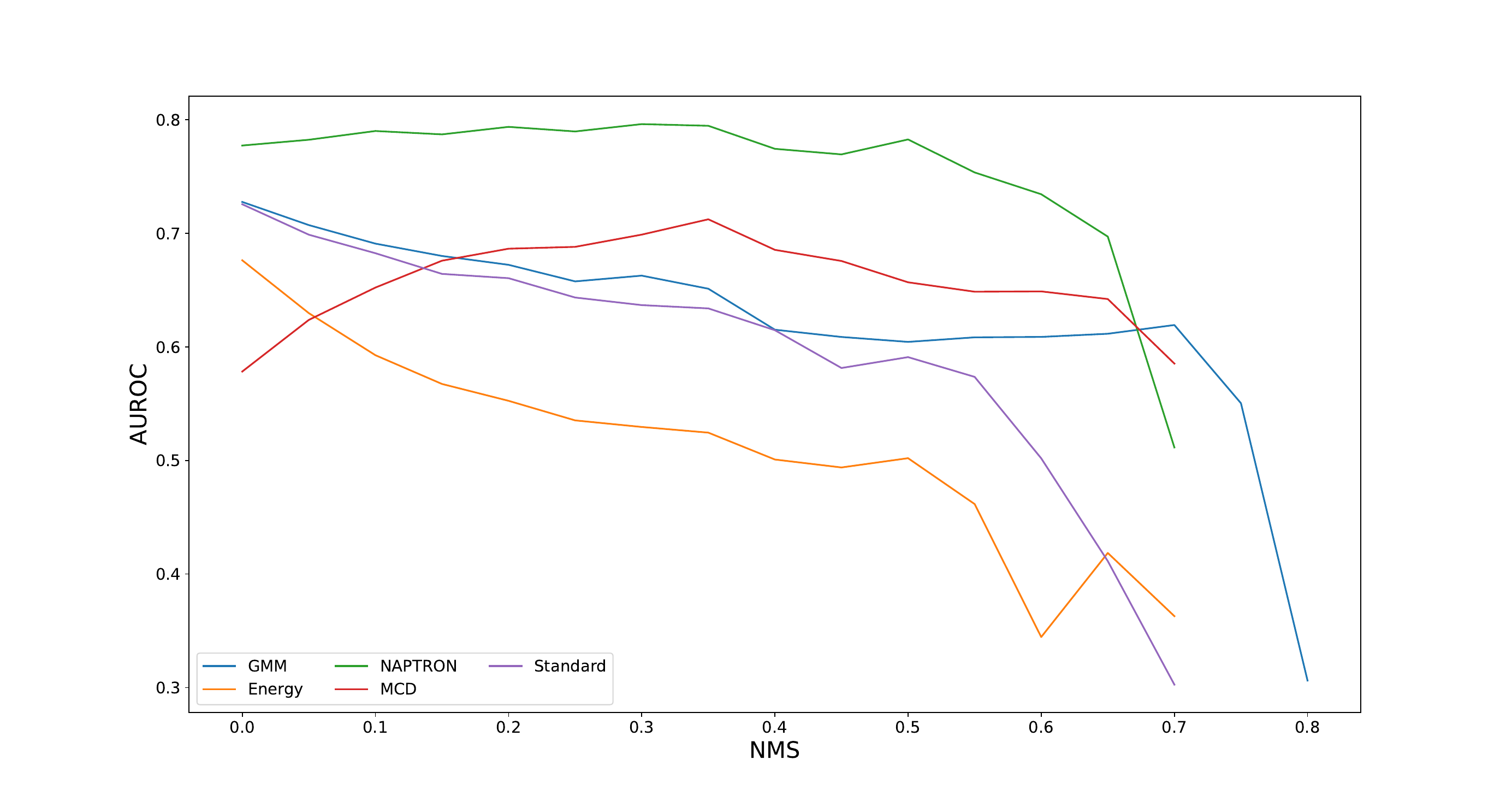}
\caption{FCOS-based OOD detectors NMS sensitivity evaluated on COCO.  Metric - AUROC.}
\label{fig:fcos-voc2coco-auroc-sensitivity}
\end{figure}

\begin{figure}[!htb]
\centering
\includegraphics[width=0.9\linewidth]{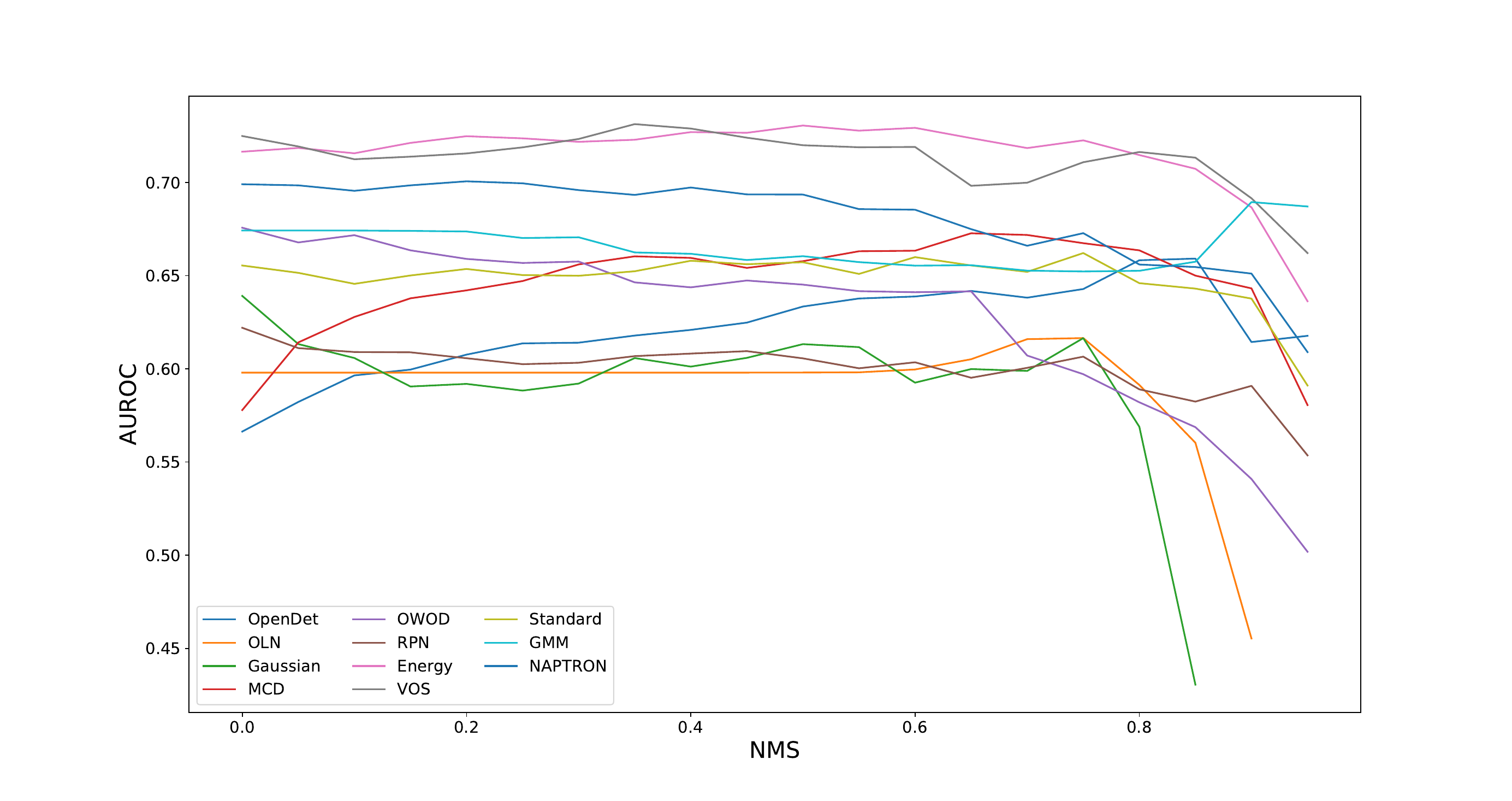}
\caption{Faster R-CNN-based OOD detectors NMS sensitivity evaluated on BDD100k. Metric - AUROC.}
\label{fig:frcnn-bdd-auroc-sensitivity}
\end{figure}

 \noindent {\bf NMS sensitivity.}
For the above experiments, we set a low NMS threshold (0.01) so that all the possible predicted objects are accounted for when computing the metrics.
However, since varying NMS softmax confidence threshold makes a significant difference in regular object detection performance, we investigate the impact of the NMS threshold on OOD detection performance. Selected NMS sensitivity plots of all considered OOD methods are presented in Fig. \ref{fig:frcnn-voc2coco-fpr-sensitivity}, \ref{fig:fcos-voc2coco-auroc-sensitivity} and \ref{fig:frcnn-bdd-auroc-sensitivity}. We included 4 out of 12 plots generated for each architecture, dataset scenario, and metric combination. %

All the figures confirm that the quality of all the OOD methods is sensitive to underlying object detectors' NMS threshold variations. However, the ranking of OOD detectors does not change much for different NMS values since most of the methods' curves share a common trend. Typically, performance metrics reach their upper limit around 0.5 and decline sharply when the softmax probability threshold is between 0.8 and 0.99. These sharp shifts of metrics' levels for high thresholds occur because object detectors hardly ever predict OOD objects with such a high probability, so OOD metrics are being computed for scarce data. This issue does not matter much in practice because, in real-world applications of object detectors, the NMS threshold is usually set somewhere between 0.5 and 0.8 - thus filtering most of the FPs and letting through most of the TPs.

\subsection{Object detection} \label{object-detection-experiments}

 Many of the state-of-the-art OOD detection methods require altering default components of models to obtain improved uncertainty scores. It is expected that these alterations should not impose any damage on regular object detection quality. Therefore, we performed additional experiments to evaluate the influence of the customization of standard object detectors. In other words, how modification of the original architecture affects the performance of the object detector.

\begin{figure}[!htb]
\centering
\includegraphics[width=0.9\linewidth]{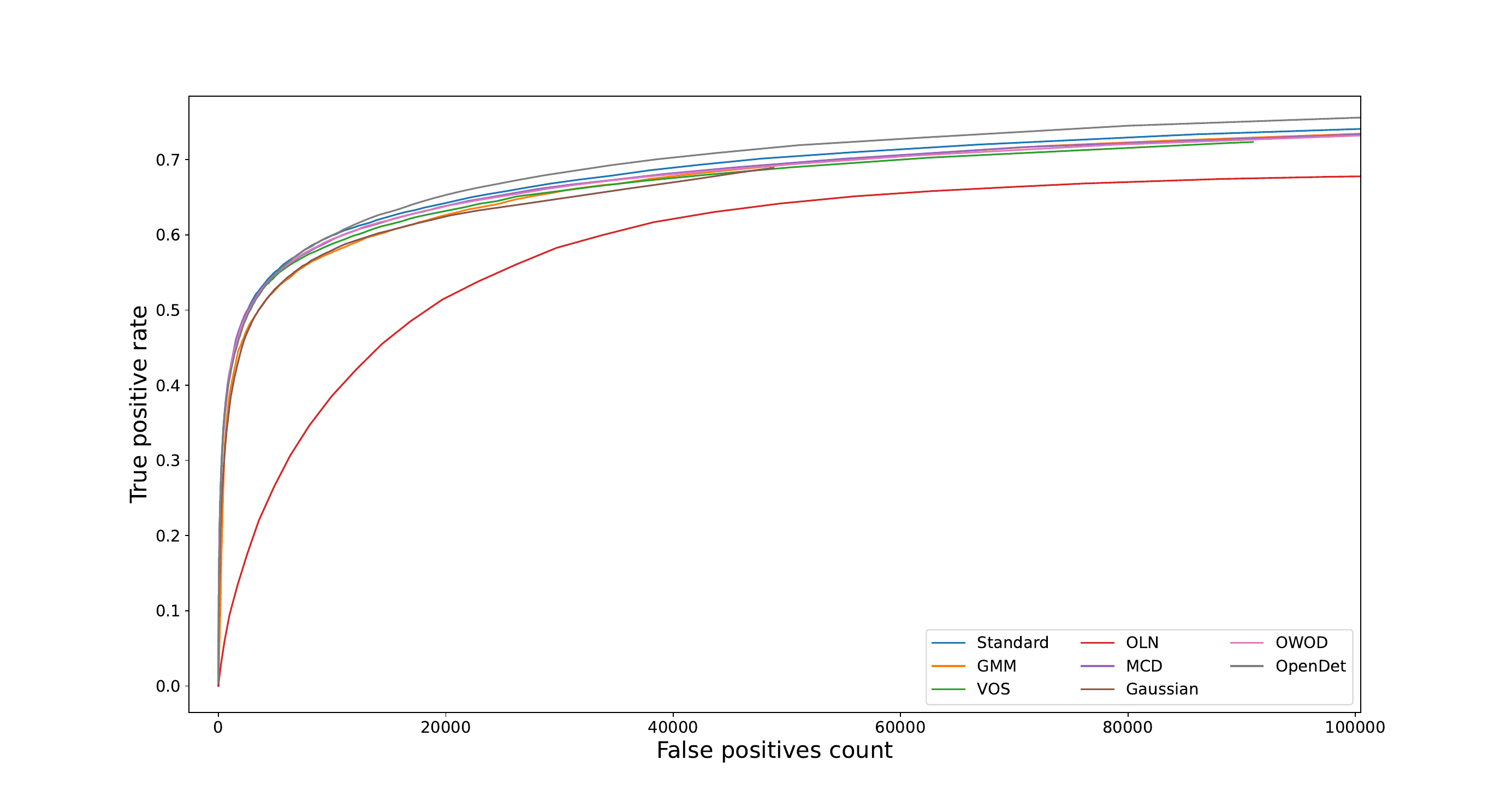}
\caption{Faster R-CNN-based detectors evaluated on COCO.}
\label{fig:frcnn-voc2coco-comparison}
\end{figure}

\begin{figure}[t]
\centering
\includegraphics[width=0.9\linewidth]{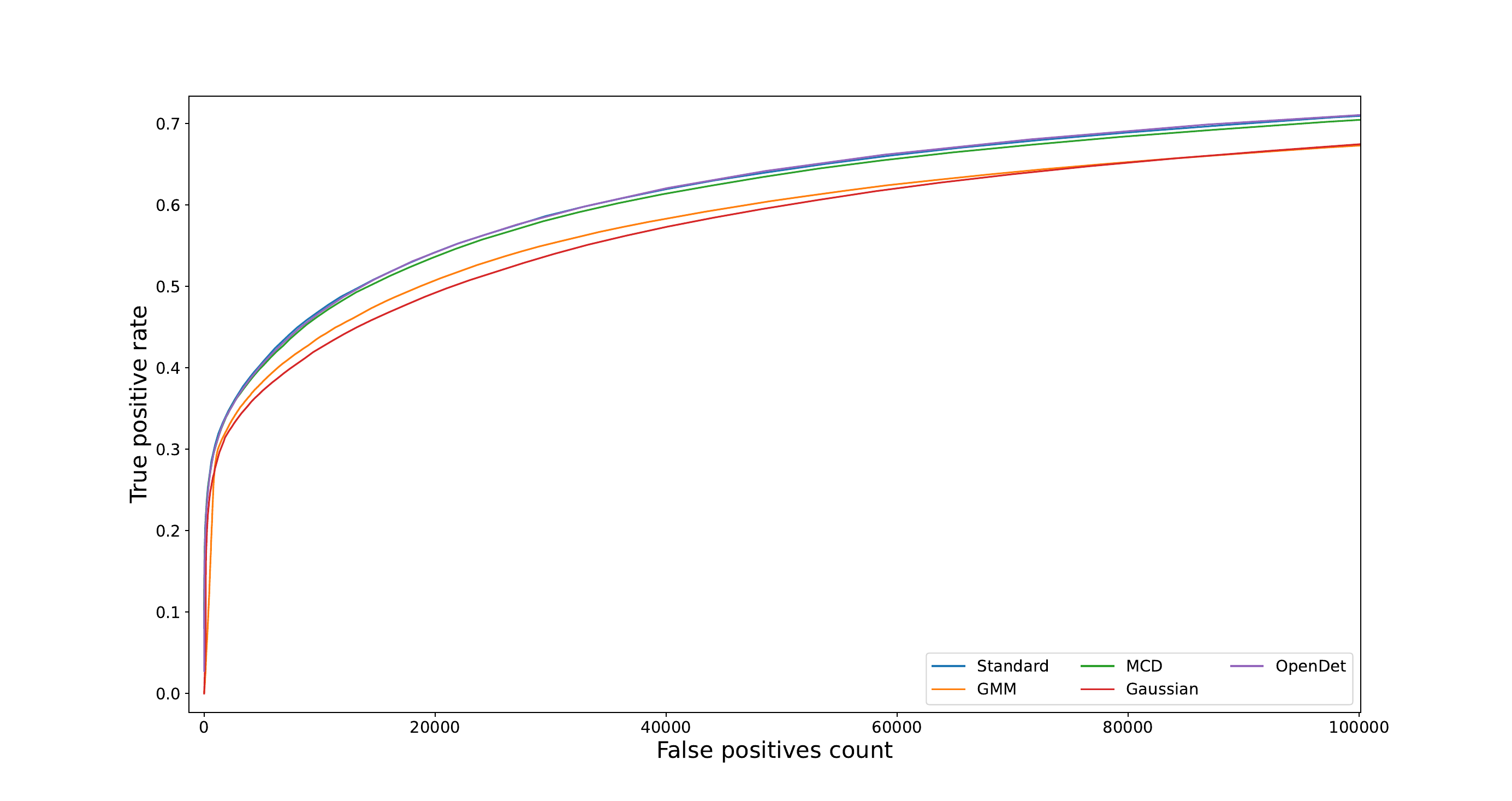}
\caption{RetinaNet-based detectors evaluated on BDD100k.}
\label{fig:retina-bdd-comparison}
\end{figure}

    Object detector performance heavily depends on a predefined, application-dependent confidence score threshold. The authors of \cite{hall2020probabilistic} observed that decreasing the NMS confidence score threshold and consequently massively increasing the number of false positive detections increases mAP. This phenomenon undermines the validity of using mAP as the primary quality metric. Therefore, as a way to gauge the impact of the OOD methods on object detection quality, we conduct a visual collation of TPR vs FP curves (see Figures \ref{fig:frcnn-voc2coco-comparison} and  \ref{fig:retina-bdd-comparison}).
  All the plots were horizontally limited to 100 000 FPs.

Drawing conclusions about the performance can hardly be completed merely by looking at the curves, so we compute the area under each curve (AUC), limiting the FP number to $2N$ where $N$ is the number of all known ground truth objects in the test dataset. Next, we compare each method's AUC with the standard detector's AUC - Table \ref{table:AUC performance} shows the results. Positive $\Delta$AUC values signify a higher (better) curve, while negative $\Delta$AUC indicates inferior performance.
 
OLN is class-agnostic (unlike all other models) and, for that reason, is subject to a more relaxed evaluation regime. The class-agnostic true positive prediction required to the correct localization of an object, without the need for accurate label assignment.

\begin{table}
\scriptsize
\begin{center}
\caption{Object detection quality - area under the curve compared to baseline.}
\label{table:AUC performance}
\begin{tabular}{llc|c} 
\hline\noalign{\smallskip}
Architecture & Method & $\Delta$AUC & $\Delta$AUC\\
\noalign{\smallskip}
\hline
\noalign{\smallskip}
& & \multicolumn{1}{c|}{VOC $\rightarrow$ COCO} & \multicolumn{1}{c}{ BDD100k  $\rightarrow$ BDD100k}  \\ \midrule
\multirow{7}{*}{Faster R-CNN} & Gaussian & -0.0214 & -0.0041 \\
& GMM & -0.0180 & -0.0015 \\
& MCD & -0.0044 & 0.0001\\
& OLN & -0.1524 & -0.1584\\
& OpenDet & 0.0073 & -0.0002  \\
& OWOD & -0.0054 & 0.0024\\
& VOS & -0.0098 & 0.0007\\
\midrule
\multirow{4}{*}{RetinaNet} & Gaussian & 0.0024 & -0.0315\\
& GMM & -0.0631 & -0.0341\\
& MCD & -0.0128 & -0.0038\\
& OpenDet & -0.0168 & 0.0019 \\
\midrule
\multirow{2}{*}{FCOS}& GMM & 0.0089 & -0.0111\\
& MCD & 0.0075 & -0.0032\\
\hline
\end{tabular}
\end{center}
\end{table}

Figures \ref{fig:frcnn-voc2coco-comparison} and \ref{fig:retina-bdd-comparison} and Table \ref{table:AUC performance} show that altering the default configurations of object detectors usually harms their performance. Except for the FCOS PASCAL-VOC $\rightarrow$ COCO that provides relatively poor results, standard detectors are the most effective. Despite more relaxed rules, OLN fails to reach the performance levels of other detectors. GMM's anchor loss proves to be detrimental, especially for RetinaNet architecture. Gaussian negative log-likelihood loss bears a negative effect on object detectors too. For a fixed number of FPs, Gaussian Faster R-CNN generates hundreds of TPs less than the standard Faster R-CNN. RetinaNet OpenDet obtains significantly worse precision in the PASCAL-VOC $\rightarrow$ COCO scenario, but in the remaining scenarios works very well. The AUC of VOS, OWOD, and MC-Dropout detectors is, on average, lower than the baseline, but the differences are smaller than for other methods. 

\subsection{Visual evaluation}


Ideally, an OOD detector provides an independent uncertainty score, yet complementary to the objects detector softmax score. By design, the softmax score is solely responsible for distinguishing background from foreground objects, and the uncertainty score separates known objects from unknown ones.

\begin{figure}[!htb]
\centering
\includegraphics[width=\linewidth]{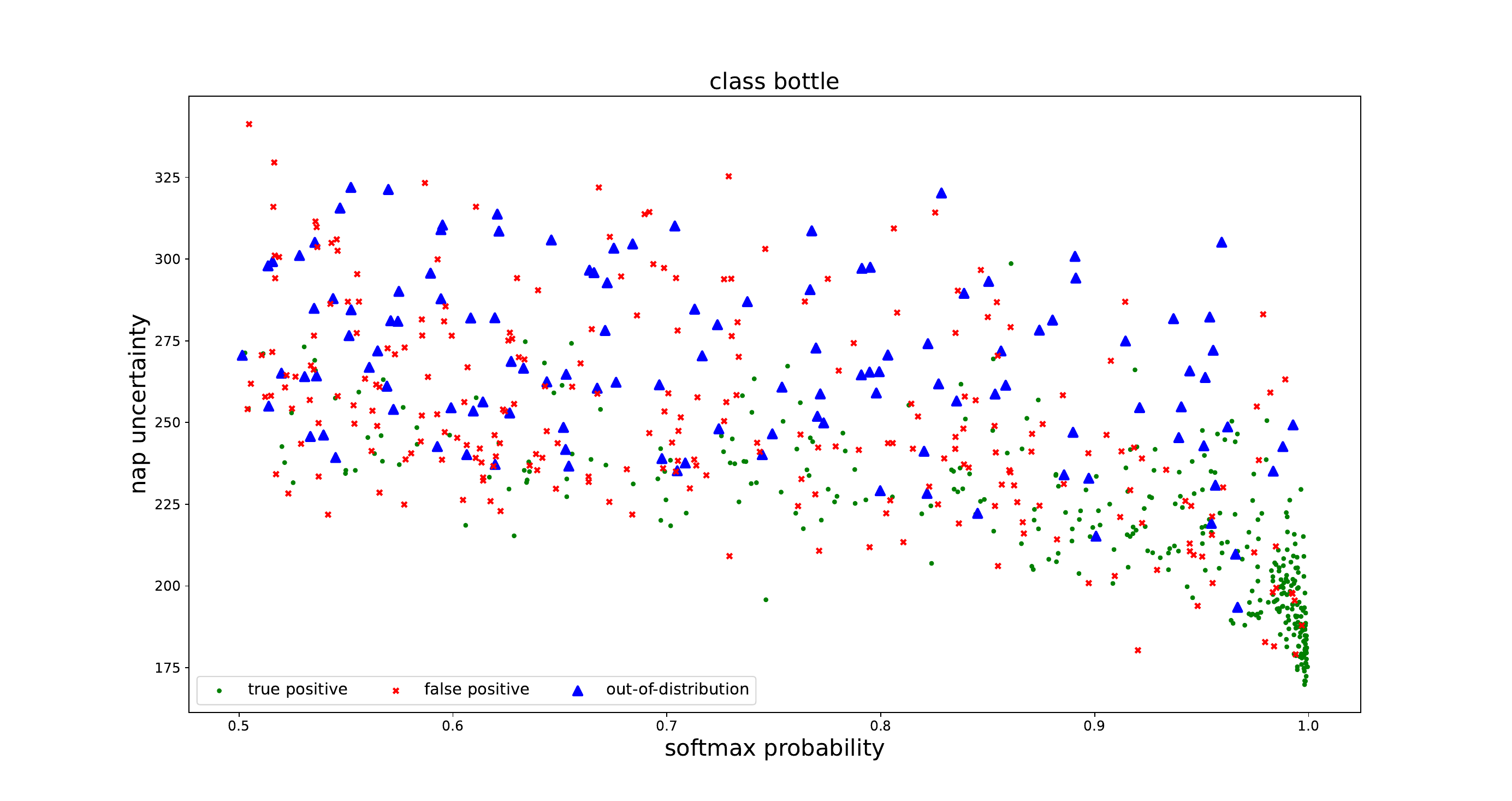}
\caption{Faster R-CNN NAPTRON bounding box uncertainty score vs softmax confidence. Class 'bottle'.}
\label{fig:frcnn-bottle-nap-uncertaintyvssoftmax}
\end{figure}

The relationship between both the scores is nontrivial and simple statistic coefficients, such as the Pearson correlation coefficient, do not explain it sufficiently. Therefore, we provide visual 2D characteristics for the best-performing OOD detectors. A perfect characteristic would depict a cloud of blue points (OOD objects) vertically separable from a cloud of green points (TPs), and a cloud of red points (FPs) horizontally separable from the TPs. 
Figure \ref{fig:frcnn-bottle-nap-uncertaintyvssoftmax} shows the NAPTRON uncertainty and softmax score relationship. We can observe that the blue OOD triangles are placed, in general, higher than the green TPs, especially for the most certain samples on the right side of the plot. 

\begin{figure}[t]

\hspace{-25mm}
\begin{subfigure}[b]{0.15\textwidth}

\includegraphics[scale=0.19, keepaspectratio]{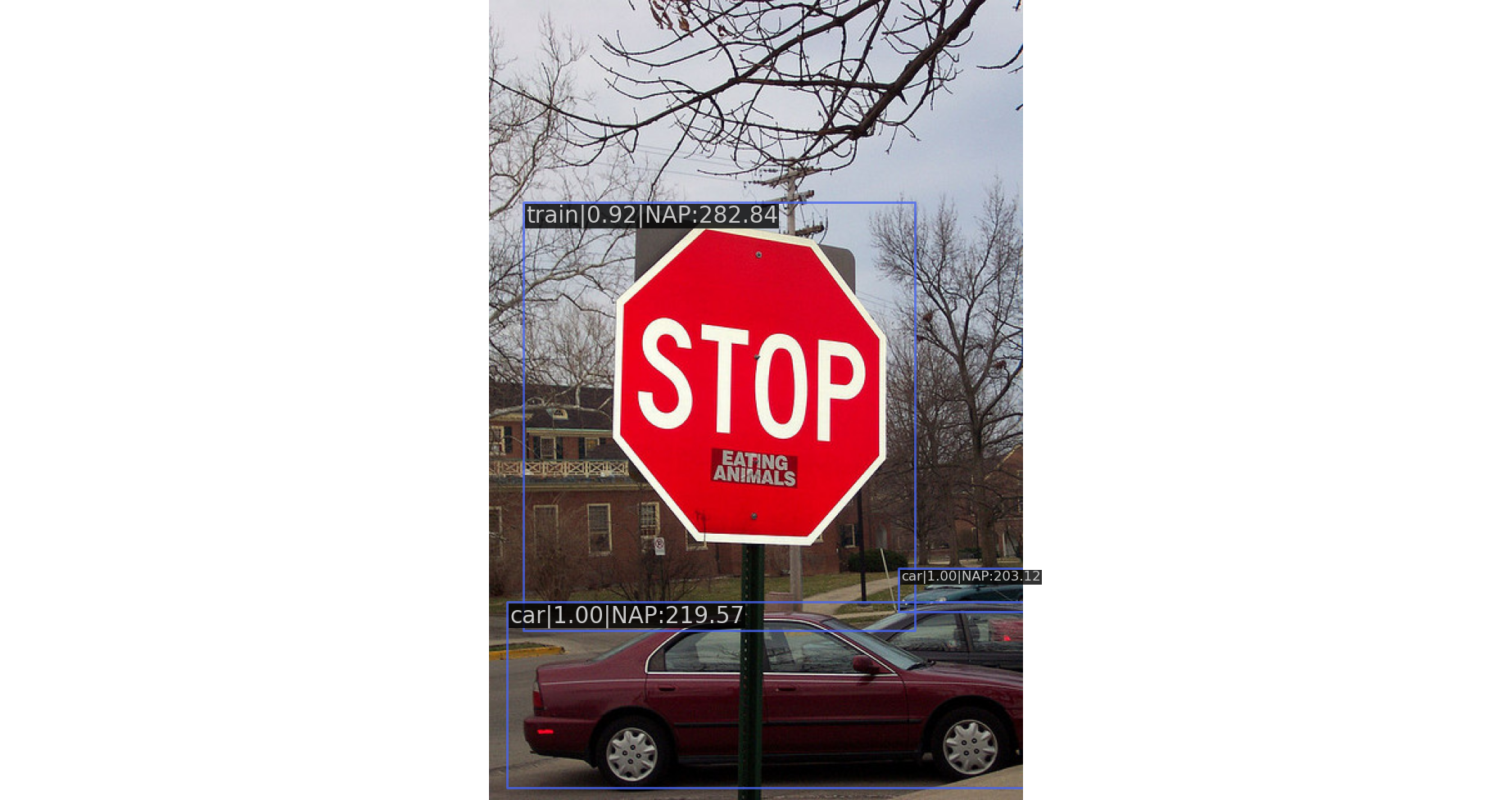}
\end{subfigure}~
\hspace{10mm}
\begin{subfigure}[b]{0.15\textwidth}
\centering
\includegraphics[scale=0.19, keepaspectratio]{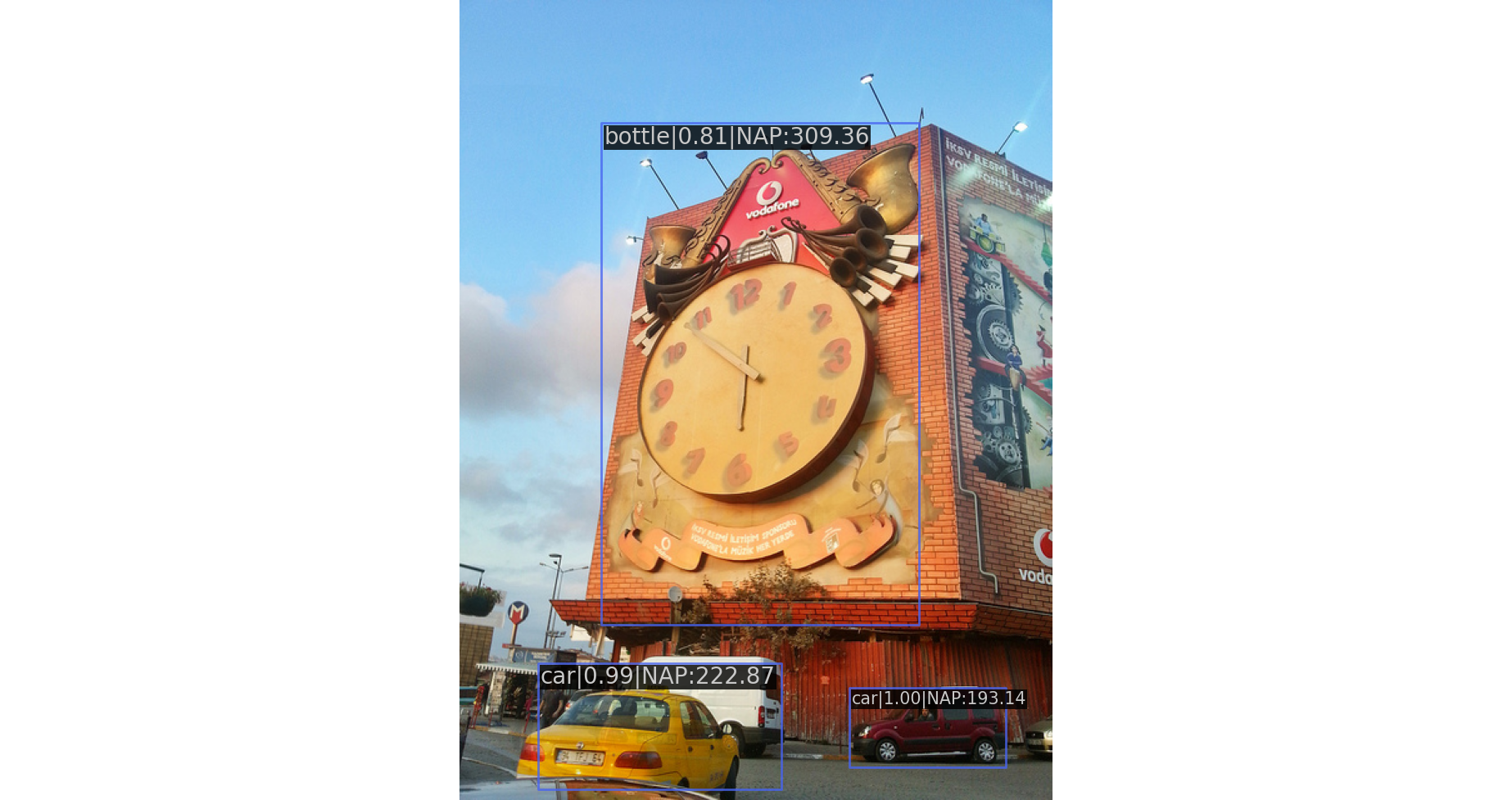}
\end{subfigure}
\caption{
1. OOD prediction - traffic sign $\rightarrow$ train 2. FP prediction background (building) $\rightarrow$ bottle.}
\label{fig:frcnn-vis1}
\end{figure}

\begin{figure}[t]
\begin{subfigure}[b]{0.42\textwidth}
\centering
\includegraphics[width=\linewidth]{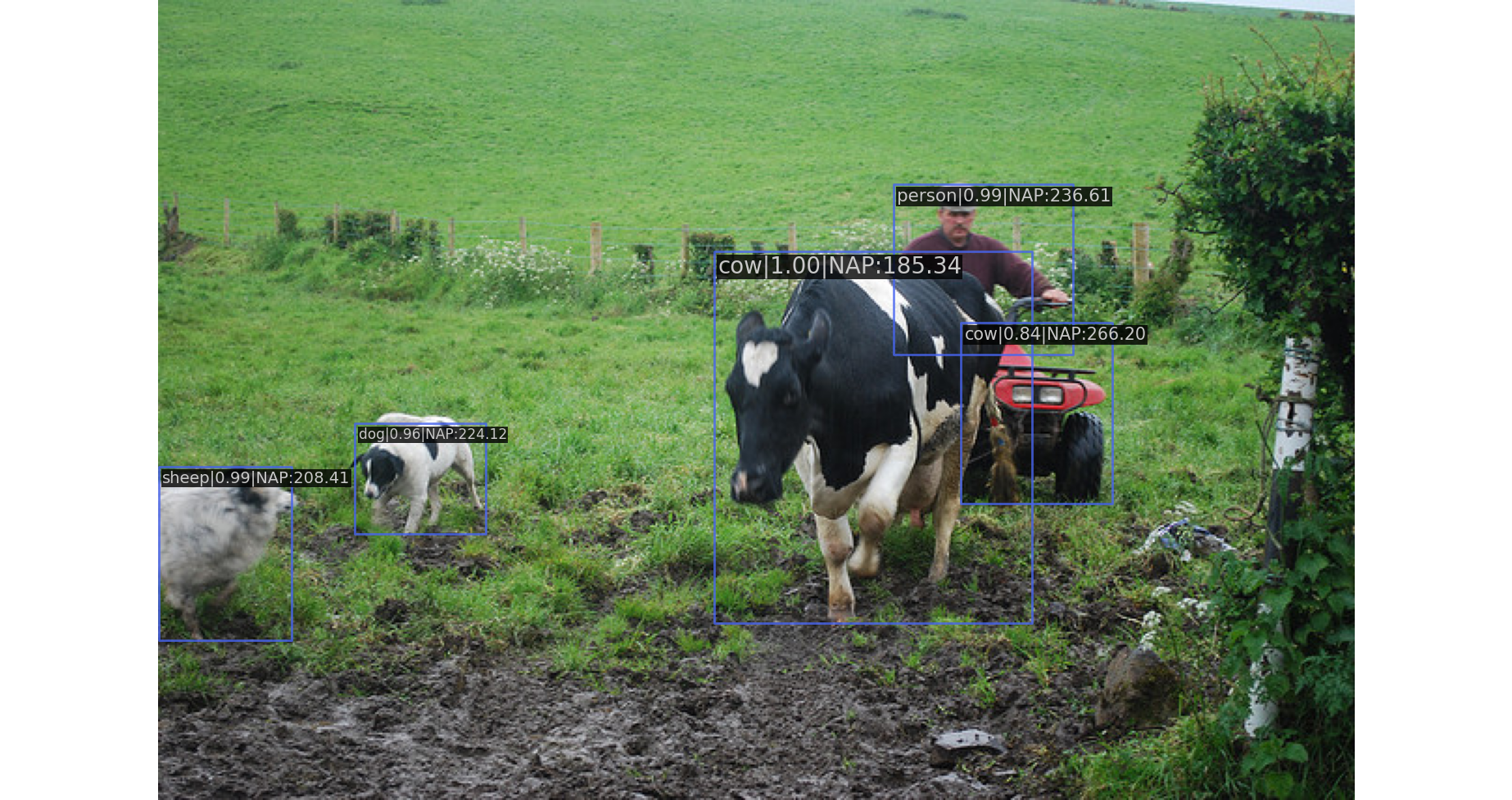}
\caption{FP prediction background (quad) $\rightarrow$ cow}
\end{subfigure}
\vskip\baselineskip
\begin{subfigure}[b]{0.42\textwidth}
\centering
\includegraphics[width=\linewidth]{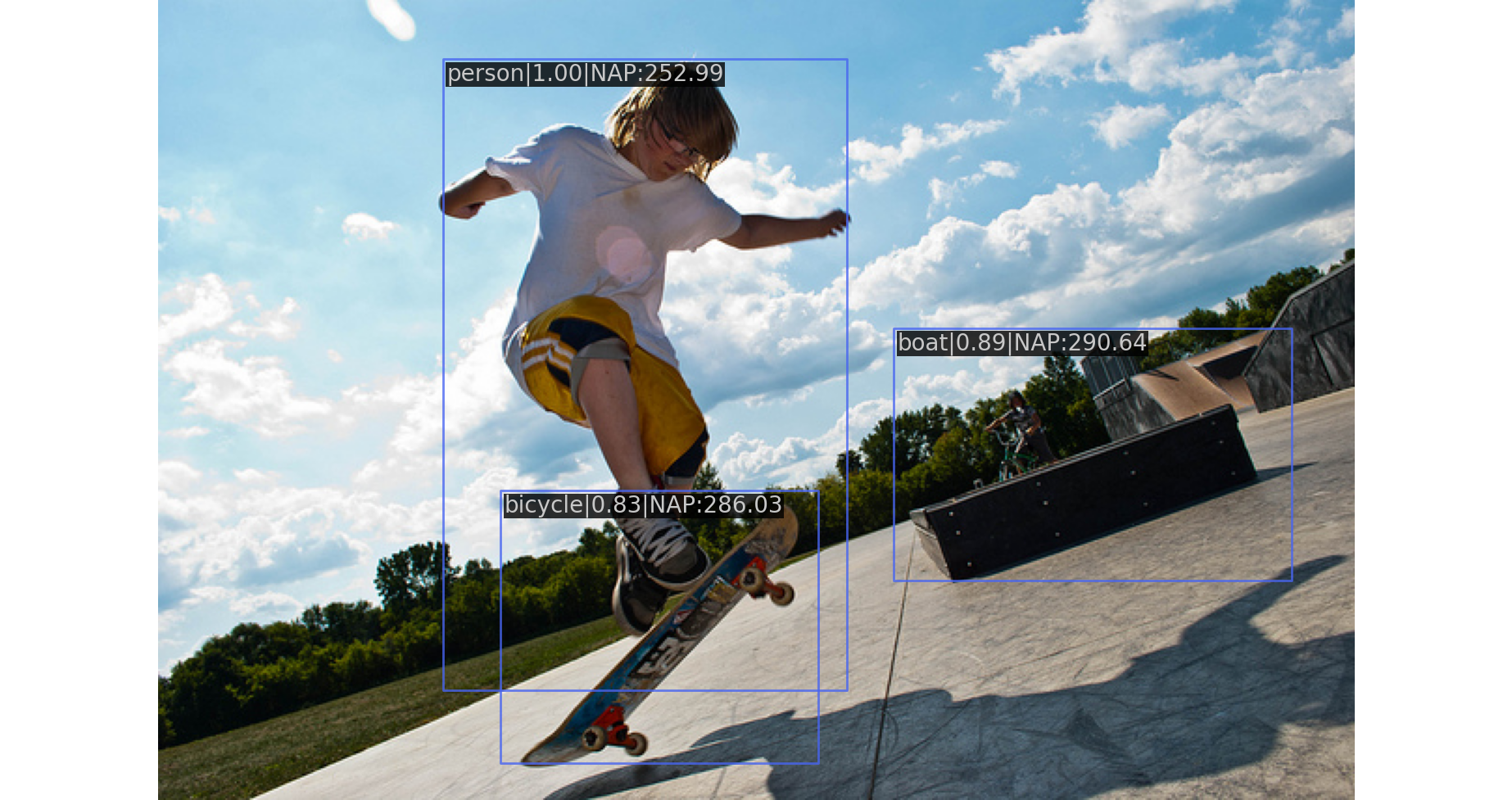}
\caption{1. OOD prediction - skateboard $\rightarrow$ bicycle 2. FP prediction background (skatepark) $\rightarrow$ boat.}
\end{subfigure}
\caption{Example of overconfident OOD and FP predictions with high NAP uncertainty. Each bounding box is attributed with softmax confidence and NAP uncertainty, respectively.}
\label{fig:frcnn-vis2}
\end{figure}

Overconfident OOD and FPs are the users' nightmare but our approach enables us to filter at least some of them. Examples of the samples identified as OOD by the proposed NAPTRON are in Figures \ref{fig:frcnn-vis1} and \ref{fig:frcnn-vis2}.
For all the examined algorithms, both scores are imperfect. In many cases, FPs have a high softmax score, whereas TPs have a lower score. Analogously, a low uncertainty score may be attributed to an OOD object. We observe that every type of uncertainty score is correlated with the softmax confidence score of an underlying object detector. This outcome is intuitive because we expect the TPs to have low uncertainty and high softmax probability. Thankfully, object detectors tend to assign lower probabilities to OOD objects, and OOD detectors assign higher uncertainty to FPs even though they are not explicitly meant to do so. These circumstances provide an opportunity to use softmax probability and one or more types of uncertainty scores combined in a manner that would boost both regular object detection and OOD detection performance.  Finding an optimal way to combine multiple scores would require another set of experiments; this is beyond the scope of this work.
    


\section{Conclusion}

In this work,  we introduce the NAPTRON algorithm to estimate the uncertainty of predicted bounding boxes, based on the analysis of binary activation patterns. Our method enables the identification of the outlying test samples by computing the Hamming distance between the sample's binary NAP and the most similar training NAP. Not only is our approach simple and intuitive but also outperforms all existing state-of-the-art OOD detectors. NAPTRON does not affect the underlying object detector's training or inference processes, which is another vital asset since we experimentally showed that altering default setups tends to lower object detection quality.  Our experiments are conducted for three different architectures representing all common kinds of object detectors, which proves the universality of our approach. 


\section*{Acknowledgements}
\noindent This work was supported by the National Centre for Research and Development under the project LIDER/51/0221/L-11/19/NCBR/2020.


\clearpage
\bibliography{ecai}
\end{document}